\documentclass[10pt,twocolumn,letterpaper]{article}
\usepackage[accsupp]{axessibility}

\usepackage{iccv}
\usepackage{times}
\usepackage{epsfig}
\usepackage{graphicx}
\usepackage{amsmath}
\usepackage{amssymb}

\usepackage{xcolor}
\usepackage{color}
\usepackage{amsmath}
\usepackage{amsfonts}
\usepackage{algpseudocode}
\usepackage{comment}

\usepackage[ruled,vlined,linesnumbered]{algorithm2e}
\usepackage{booktabs} 
\usepackage{multirow,makecell}

\usepackage[para,online,flushleft]{threeparttable}
\usepackage[switch]{lineno}
\usepackage[caption=false]{subfig}
\usepackage{multicol}

\newcommand{\hw}[1]{\ensuremath{\mathtt{#1}}}

\usepackage[pagebackref=true,breaklinks=true,letterpaper=true,colorlinks,bookmarks=false]{hyperref}

\iccvfinalcopy 

\def\iccvPaperID{7280} 

\ificcvfinal\fi

\begin{document}

\title{RMSMP: A Novel Deep Neural Network Quantization Framework with Row-wise Mixed Schemes and Multiple Precisions}

\author{Sung-En Chang$^{*1}$, Yanyu Li$^{*1}$, Mengshu Sun$^{*1}$, \thanks{Equal contribution. }Weiwen Jiang$^{2}$, Sijia Liu$^{3}$, Yanzhi Wang$^{1}$, Xue Lin$^{1}$\\

$^1$Northeastern University, $^2$George Mason University, $^3$Michigan State University\\
{\tt\small \{chang.sun, li.yanyu, sun.meng, yanz.wang, xue.lin\}@northeastern.edu,}\\
{\tt\small wjiang8@gmu.edu, liusiji5@msu.edu}
}

\maketitle
\ificcvfinal\thispagestyle{empty}\fi

\begin{abstract}
This work proposes a novel Deep Neural Network (DNN) quantization framework, namely RMSMP, with a \underline{R}ow-wise \underline{M}ixed-\underline{S}cheme and \underline{M}ulti-\underline{P}recision approach. Specifically, this is the first effort to assign mixed quantization schemes and multiple precisions within layers -- among rows of the DNN weight matrix, for simplified operations in hardware inference, while preserving accuracy. Furthermore, this paper makes a different observation from the prior work that the quantization error does not necessarily exhibit the layer-wise sensitivity, and actually can be mitigated as long as a certain portion of the weights in every layer are in higher precisions. This observation enables layer-wise uniformality in the hardware implementation towards guaranteed inference acceleration, while still enjoying row-wise flexibility of mixed schemes and multiple precisions to boost accuracy. The candidates of schemes and precisions are derived practically and effectively with a highly hardware-informative strategy to reduce the problem search space. 

With the offline determined ratio of different quantization schemes and precisions for all the layers, the RMSMP quantization algorithm uses Hessian and variance based method to effectively assign schemes and precisions for each row. The proposed RMSMP is tested for the image classification and natural language processing (BERT) applications, and achieves the best accuracy performance among state-of-the-arts under the same equivalent precisions. The RMSMP is implemented on FPGA devices, achieving $3.65\times$ speedup in the end-to-end inference time for ResNet-18 on ImageNet, comparing with the 4-bit Fixed-point baseline.
\end{abstract}


\section{Introduction}

With tremendous success of the deep learning or deep neural networks (DNNs), there exist urgent needs for deployments of inference models onto edge-computing platforms/devices.
As a major type of model compression technique, the DNN quantization becomes an essential method to reduce the computation, memory, and storage requirements in on-device inference, especially for platforms with capability of customized architecture design, such as FPGA devices and ASIC chips.
Generally speaking, DNN quantization learns DNN models in low bit-width representation with accuracy performance close to that of the full-precision models, while accelerating inference speed.

\begin{figure}[b]
\centering  
\includegraphics[width=1.0\columnwidth]{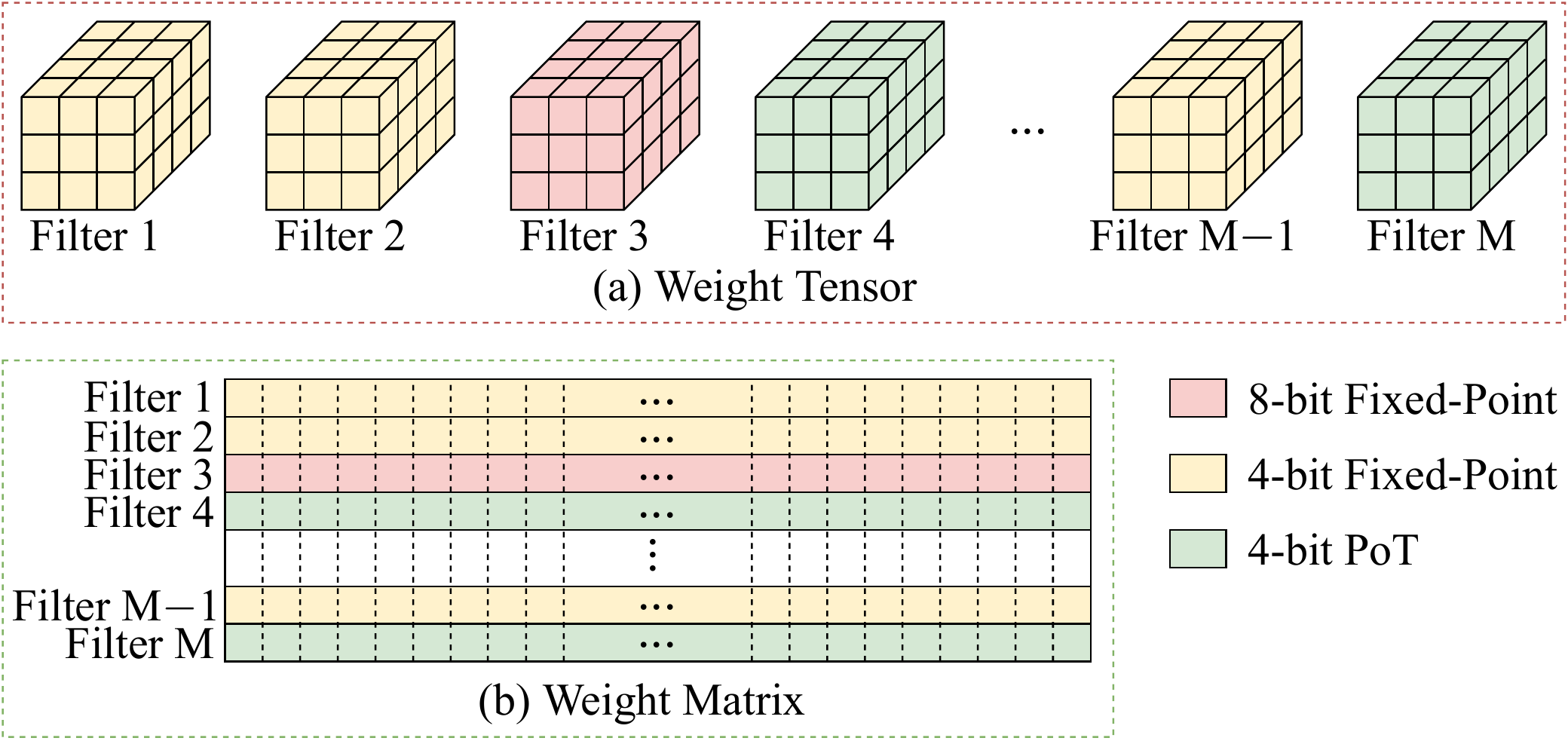}
\caption{\textbf{The proposed DNN quantization framework with row-wise mixed schemes and multiple precisions, which assigns quantization scheme and precision to filters of the weight tensor (or rows of the weight matrix).} It features (i) layer-wise uniformality to fulfill the requirement of practical hardware implementation, (ii) row-wise flexibility for mixed schemes and multiple precisions, (iii) hardware-informative selection of candidate schemes and precisions (bit-widths) for significantly reducing the algorithm search space, and (iv) superior accuracy performance among the state-of-the-arts.}
\label{fig:IntraLayer}
\end{figure}

Various quantization schemes have been investigated including binary~\cite{courbariaux2015binaryconnect,courbariaux2016binarized,rastegari2016xnor,lin2017towards}, ternary\cite{li2016ternary,he2019simultaneously,zhu2016trained}, Fixed-point (Fixed)~\cite{zhou2016dorefa,choi2018pact,gong2019differentiable,jung2019learning,cheng2019uL2Q,esser2019learned}, Power-of-Two (PoT)~\cite{DBLP:journals/corr/MiyashitaLM16,DBLP:journals/corr/ZhouYGXC17,leng2018extremely,zhang2018lq}, Additive Power-of-Two (APoT)~\cite{li2019additive}, etc.
Those schemes have diverse accuracy and hardware performance.
Binary and ternary significantly reduce computation by eliminating multiplication operations, but experience relatively large accuracy loss ($>2\%$ in general).
Low bit-width Fixed quantization has better accuracy performance. 
For example, 4-bit Fixed can achieve negligible accuracy loss comparing with its 32-bit floating-point counterpart, although it still needs multiplication operations during inference computation.

Different from binary, ternary, and Fixed, PoT is a non-linear quantization scheme, where with quantization levels as power-of-two numbers, multiplications can be replaced with bit shifting operations, thereby reducing computation to speedup  inference.
However, PoT still results in moderate accuracy loss ($1\%\sim2\%$), which is due to the rigid resolution issue \cite{li2019additive} that exhibits a high resolution around the mean and a low resolution at the tails of weight distribution and that cannot be resolved even with higher bit-width.
Therefore, APoT was proposed by representing quantization levels as the sum of multiple power-of-two numbers, overcoming the rigid resolution issue of PoT, while enjoying non-multiplication operations.

While exploring various quantization schemes, people found that the first and the last layers are of importance for preserving accuracy, and therefore in the above-mentioned works, the first and the last layers are either unquantized (in 32-bit floating-point) or quantized with no fewer than 8 bits.
Motivated by this, the layer-wise multi-precision quantization has been well investigated in \cite{wang2019haq,wu2018mixed,uhlich2019mixed,dong2019hawq,dong2019hawqv2,yao2019pyhessian,shen2020q}, where various methods/algorithms have been used to deal with the large search space of the precision assignment for both weights and activations in each layer.

This work proposes a novel DNN quantization framework, namely RMSMP, with a \underline{R}ow-wise \underline{M}ixed-\underline{S}cheme and \underline{M}ulti-\underline{P}recision approach.
Figure \ref{fig:IntraLayer} illustrates the proposed quantization framework on the DNN (a) weight tensor and (b) weight matrix.
Specifically, each filter in the weight tensor or each row in the weight matrix is assigned a combination of quantization scheme and precision. 
The candidates of schemes and precisions are derived \emph{practically} for facilitating hardware implementation, and \emph{effectively} for speeding up inference with the given resources on the hardware platforms/devices, while with the capability to preserve the accuracy as the unquantized (32-bit floating-point) models.
This highly hardware-informative quantization strategy significantly reduces the search space of the DNN quantization problem, making our framework distinctive from existing multi-precision quantization works.

This is \emph{the first effort to apply mixed quantization schemes and multiple precisions} within layers, targeting for simplified operations in hardware inference, while preserving the accuracy.
Specifically, two quantization schemes i.e., Power-of-Two (PoT) and Fixed-point (Fixed), and two precisions i.e., 4-bit and 8-bit are adopted and explored for quantization on weights and activations, to reduce inference computation and preserve accuracy.
Furthermore, as demonstrated by the previous works that either implicitly use higher precisions for the first and the last layers, or explicitly assign multiple precisions to different layers, it is essential to use higher precisions at least for parts of the model, to boost the accuracy of quantized models close to that of the full-precision models.
However, different from existing works, this paper makes the observation that this does not necessarily relate to layer-wise sensitivity, instead, as long as a certain portion of the weights in every layer use higher precisions, the quantization error can be mitigated.
This observation enables layer-wise uniformality towards practical hardware implementation of quantized models, while still enjoying row-wise flexibility of mixed schemes and multiple precision.
The contributions of our quantization framework are summarized as follows:
\begin{itemize}
    \item \textbf{A novel row-wise mixed-scheme and multiple-precision quantization approach.}
    
    \item \textbf{A highly hardware-informative solution strategy significantly reducing the problem search space.} 
    
    \item \textbf{The best accuracy performance when under the same equivalent precision as the existing works.}
    
    \item \textbf{The significant inference speedup on real devices, comparing to the network-wise uniform low-bit quantization i.e., the speed upper bound of the popular layer-wise multi-precision approaches.}

\end{itemize}

\section{Related Work}\label{sec:relatedwork}


\subsection{Quantization Scheme}\label{sec:quantizationscheme}


\subsubsection{Linear Quantization Schemes}

Linear quantization schemes with uniform quantization levels include binary, ternary, and fixed-point.
Both binary and ternary quantization use extremely low precision for DNN models, i.e., binarized (with values of $-1, +1$) or ternarized (with values of $-1, 0, +1$) levels.
Representative binary quantization methods include Binaryconnect~\cite{courbariaux2015binaryconnect}, Binarized Neural Network (BNN)~\cite{courbariaux2016binarized}, XNOR-net~\cite{rastegari2016xnor}, and ABC-Net~\cite{lin2017towards}.
With weights constrained to $\{-1, +1\}$, multiplications can be replaced by additions/subtractions, which can even be eliminated by using XNOR and AND operations if activations are quantized to binary as well. 
Ternary quantization schemes are implemented in TWN~\cite{li2016ternary}, TTQ~\cite{zhu2016trained}, and~\cite{he2019simultaneously}. Ternary networks also benefit from non-multiplication operations. 
Although binary and ternary quantization can significantly reduce computation by simplifying operations, they introduce relatively large accuracy loss. From the above studies, accuracy typically degrades by $>5\%$ under binary, and $2\% \sim 3\%$ for ternary.

Fixed-point (Fixed) quantization schemes use more bits than binary and ternary to preserve accuracy, and have been implemented with different methods/algorithms.
DoReFa-Net~\cite{zhou2016dorefa} first explored it by introducing hyperbolic tangent transformation to weights and activations, with scaling factors to minimize quantization error. PACT~\cite{choi2018pact}  improved this method by adding a parameterized clipping threshold to activations. DSQ~\cite{gong2019differentiable} 
developed differentiable soft quantization, which evolves training method to gradually approximate the uniform quantizer.
QIL~\cite{jung2019learning} parameterized the quantization interval and trained it with task loss, avoiding access to the original training data. $\mu$L2Q~\cite{cheng2019uL2Q} introduced data distribution loss during training to minimize quantization error. LSQ~\cite{esser2019learned} proposed a differentiable method to learn the quantizer for each layer jointly with parameters.
Generally, 4-bit fixed-point quantized models have negligible accuracy loss comparing with 32-bit floating-point models, although cannot get rid of the expensive multiplications. 

\subsubsection{Non-Linear Quantization Schemes}

Non-linear quantization schemes use non-uniform quantization levels, such as the Power-of-Two (PoT) scheme, where quantization levels are power-of-two numbers.
The multiplication of a weight (in PoT) and an input (in Fixed) can be replaced with the bit shifting operation. 
And therefore PoT can have higher speedup than the Fixed scheme for the DNN inference.
PoT quantization was adopted first in LogQuant~\cite{DBLP:journals/corr/MiyashitaLM16}, and then with model accuracy enhancement techniques in INQ~\cite{DBLP:journals/corr/ZhouYGXC17}, extremely low bit neural network with ADMM~\cite{leng2018extremely}, and LQ-Nets~\cite{zhang2018lq}.
However, PoT suffers from the rigid resolution phenomenon, and cannot attain higher model accuracy even with increased bit-width. 
Particularly, 4-bit PoT quantization will result in accuracy loss of $1\% \sim 2\%$.

To overcome the limitation, Additive Power-of-Two (APoT)~\cite{li2019additive} represents quantization levels as the sum of multiple power-of-two numbers to mitigate the accuracy loss. 
Then the multiplication of an APoT quantized weight and a Fixed input can be replaced with bit shifting operations and additions.
Furthermore, MSQ~\cite{chang2021mix} leverages a mixture of a variant APoT and Fixed to maximize the resource utilization on FPGA devices, while in this work, we use mixed schemes of PoT and Fixed, achieving even higher accuracy and further reducing inference computation. 
All of the above works use the single-precision quantization.

\subsection{Single-Precision vs Multi-Precision}

A common practice in the above-mentioned quantization works is to keep the first and last layers unquantized (in 32-bit floating-pointing for weights/activations i.e., W32A32) or quantized with no fewer than 8 bits, with the purpose of preserving accuracy, which was first proposed in \cite{han2015learning}. 
Spurred by that, another line of works ~\cite{wang2019haq,wu2018mixed,uhlich2019mixed,dong2019hawq,dong2019hawqv2,yao2019pyhessian,shen2020q} explore the layer-wise multi-precision quantization approach to boost the accuracy of quantized models.

To deal with the large search space of layer-wise precision assignment for both weights and activations, different methods/algorithms have been leveraged.
HAQ~\cite{wang2019haq} uses the reinforcement learning to train an agent that decides the bit-width of each layer. 
DNAS~\cite{wu2018mixed} proposes a multi-precision quantization-aware training technique, where the bit-width search is converted into a network architecture search. 
Mixed Precision DNNs~\cite{uhlich2019mixed} shows that bit-widths can be determined by learnable parameters, whose gradients are estimated using Straight Through Estimator (STE). 
HAWQ~\cite{dong2019hawq}, HAWQ-V2~\cite{dong2019hawqv2}, and PyHessian~\cite{yao2019pyhessian} solve Hessian matrix to determine bit-width for each layer. 
The general idea is to assign more bits to layers that are sensitive to quantization error. 
Q-BERT~\cite{shen2020q} also uses Hessian-based multi-precision quantization to the BERT ~\cite{devlin2018bert} model for the natural language processing application.

There are two frameworks applying  multi-precision within layers. 
RVQuant~\cite{park2018value} quantizes a small percentage of weights with a higher precision irregularly within layers, and therefore could not achieve inference speedup comparing with a network-wise uniform  low-bit quantization.
AutoQ~\cite{lou2019autoq} trains a reinforcement learning agent to assign precisions down to the kernel level, which is computationally expensive. 
None of them considers mixed schemes.

\section{Proposed RMSMP Quantization}

This section discusses the proposed RMSMP quantization framework with a highly hardware-informative strategy, by introducing the scheme selection, precision exploration, and the algorithm solution.

\subsection{Mixed Schemes for Simplified Operations}

The Fixed-point (Fixed) quantization scheme has superior accuracy performance, and the Power-of-Two (PoT) is the most computationally efficient quantization scheme (with still acceptable accuracy performance) to speedup inference since multiplications can be replaced by bit shifting operations.
Therefore, this work proposes a novel row-wise mixed-scheme quantization approach with Fixed for preserving accuracy and PoT for reducing computation of inference.
Specifically, for each row in the weight matrix or each filter in the weight tensor, the weights are either quantized into the Fixed scheme or the PoT scheme.
The row-wise scheme assignment instead of a layer-wise scheme assignment is used, because the hardware inference execution is conducted layer by layer on the same pieces of computing resource -- \hw{GEMM_{Fixed}} i.e., the GEMM (general matrix multiply) core for processing Fixed weights, and \hw{GEMM_{PoT}} i.e., the GEMM core for processing PoT weights. 
(Note that in PoT, activations are also quantized into Fixed to support the bit shifting operation in replacement of multiplication.)

The ratio of the two schemes is fixed based on the hardware characteristics, and is the same for different layers to keep the layer-wise uniformality in hardware design such that inference speedup can be obtained in the layer-by-layer inference execution.
In ASIC chips, it is desirable to maximize the portion of the PoT quantized rows to the extent that it does not result in accuracy loss.
In FPGA devices, there are heterogeneous computing resources i.e., DSPs to implement the \hw{GEMM_{fixed}} core and LUTs to implement the \hw{GEMM_{PoT}} core.
Therefore, the ratio of the two schemes could be OFFLINE determined, such that the LUT utilization is high (while with 100\% DSP utilization) to speed up inference and also it should not hurt accuracy.

In the training algorithm of our RMSMP quantization, the row-wise scheme assignment is based on the weight distribution of the particular row.
If the weight distribution has a smaller variance, the PoT scheme is preferred; otherwise, the Fixed scheme is preferred.
The threshold on the variance can be used based on the offline determined ratio of the two schemes.

\subsubsection{Fixed-Point (Fixed) Quantizer}
The Fixed scheme uses uniform quantization levels, with the quantized weights mapped as a scaling factor times the quantization levels.
With $m$-bit Fixed, the quantized weights should be from the following set:
\begin{equation}\label{eq:fixedpointQL}
\scalebox{1.0}{
\begin{math}
\displaystyle
\mathcal{Q}^\text{Fixed}(m, \alpha) = \pm\alpha \times \{0, \frac{1}{2^{m-1}-1}, \frac{2}{2^{m-1}-1}, \dots,  1\}.
\end{math}
}
\end{equation}
where $\alpha$ denotes the scaling factor. 
Then the quantizer function from a 32-bit floating-point weight $w$ to a Fixed quantized weight $\hat w$ is
\begin{equation}\label{eq:fixedpointquantizer}
\scalebox{1.0}{
\begin{math}
\displaystyle
\begin{aligned}
\hat w & = \prod_{\mathcal{Q}^\text{Fixed}(m, \alpha)} w \\
& = \alpha\cdot h^{-1}\bigg(\frac{1}{2^{m}-1} \cdot \text{round} \big( (2^{m}-1) \cdot h \big( \lceil w,\alpha\rfloor \big) \big) \bigg),
\end{aligned}
\end{math}
}
\end{equation}
where $h(\cdot)$ shifts a value within $[-1,+1]$ into the range of $[0,1]$, e.g., $h(x)=x/2+0.5$, and $\lceil w,\alpha\rfloor$ clips $w$ according to
\begin{equation}
    \begin{aligned}
    \lceil w,\alpha\rfloor = 
    \begin{cases}
    -1, \quad &w<-\alpha\\
    w/\alpha, \quad &-\alpha\leq w \leq\alpha\\
    1, \quad & w>\alpha
    \end{cases}
    \end{aligned}
.\end{equation}

\subsubsection{Power-of-Two (PoT) Quantizer}

The PoT scheme uses power-of-two numbers as quantization levels.
With $m$-bit PoT, the quantized weights should be from the following set:
\begin{equation}\label{eq:P2QL}
\mathcal{Q}^\text{PoT}(m, \alpha)= \pm\alpha\times \{0, \frac{1}{2^{2^{m-1}-2}}, \frac{1}{2^{2^{m-1}-3}}, \dots, 1\}
.\end{equation}
Then the quantizer function from a 32-bit floating point weight $w$ to a PoT quantized weight $\hat w$ is
\begin{equation}\label{eq:powerof2quantizer}
\begin{aligned}
\hat w
& = \prod_{\mathcal{Q}^\text{PoT}(m, \alpha)} w\\
& =\begin{cases}
\begin{aligned}
&\alpha \cdot h^{-1}\big(2^{\text{round}(\log_2 h')}\big), & h'>2^{-2^m+1} \\
&0, & h'\leq 2^{-2^m+1} \\
\end{aligned}
\end{cases},
\end{aligned}
\end{equation}
where $h'= h(\lceil w,\alpha\rfloor)$.

\subsection{Multiple Precisions for Boosting Accuracy}

\begin{figure}[b]
\centering  
\includegraphics[width=0.6\columnwidth]{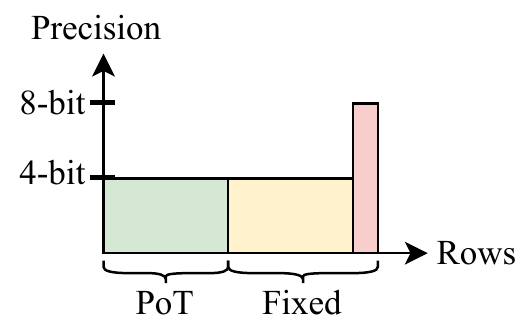}
\caption{\textbf{The row-wise mixed-scheme and multiple-precision quantization.} The PoT scheme uses only the 4-bit precision for weights/activations i.e., PoT-W4A4, and the Fixed scheme uses two precisions i.e., Fixed-W4A4 and Fixed-W8A4.}
\label{fig:Precision}
\end{figure}

Motivated by the previous work that either implicitly use higher precisions for the first and the last layers, or explicitly assign multiple precisions to different layers, it is essential to use higher precisions for some parts of the model to boost the accuracy.
However, the layer-wise multi-precision approach does not compatible with the layer-by-layer inference execution on the same pieces of hardware.
In other words, the layer-wise multi-precision approach incurs large implementation overhead, which may neutralize the expected inference speedup.

\begin{figure}[t]
\centering  
\includegraphics[width=0.85\columnwidth]{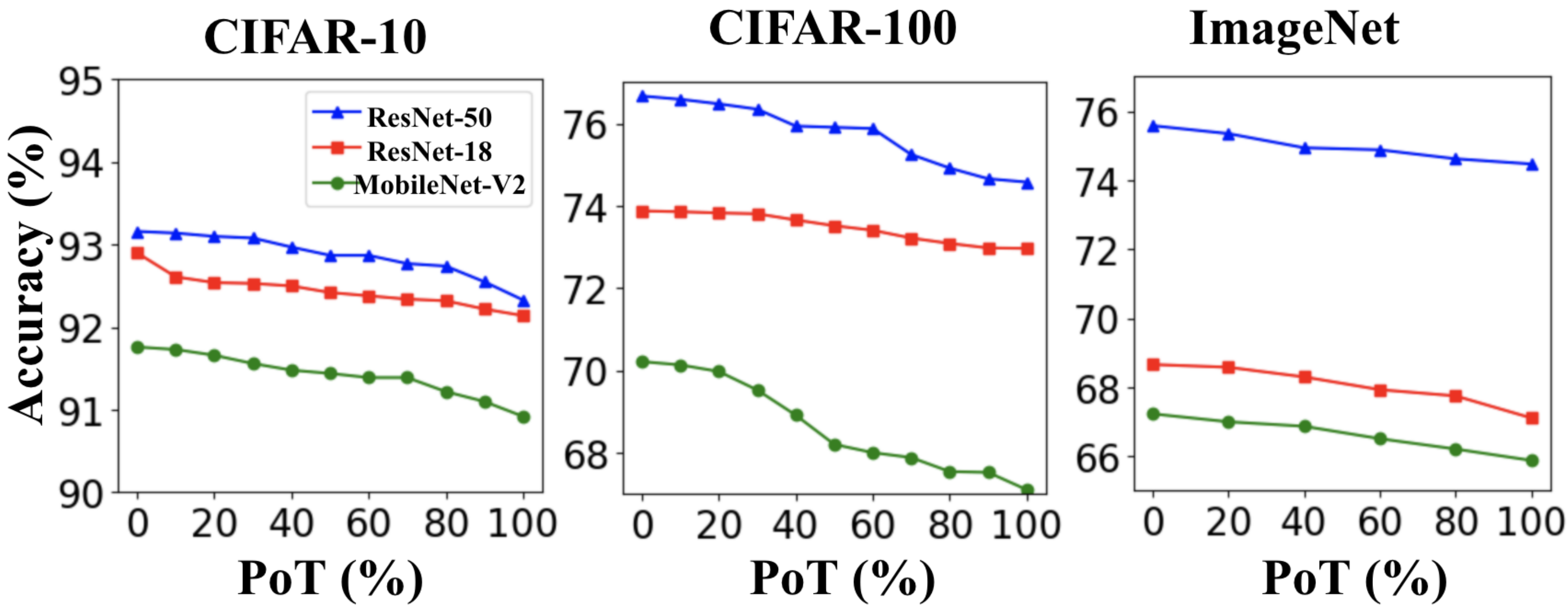}
\includegraphics[width=0.85\columnwidth]{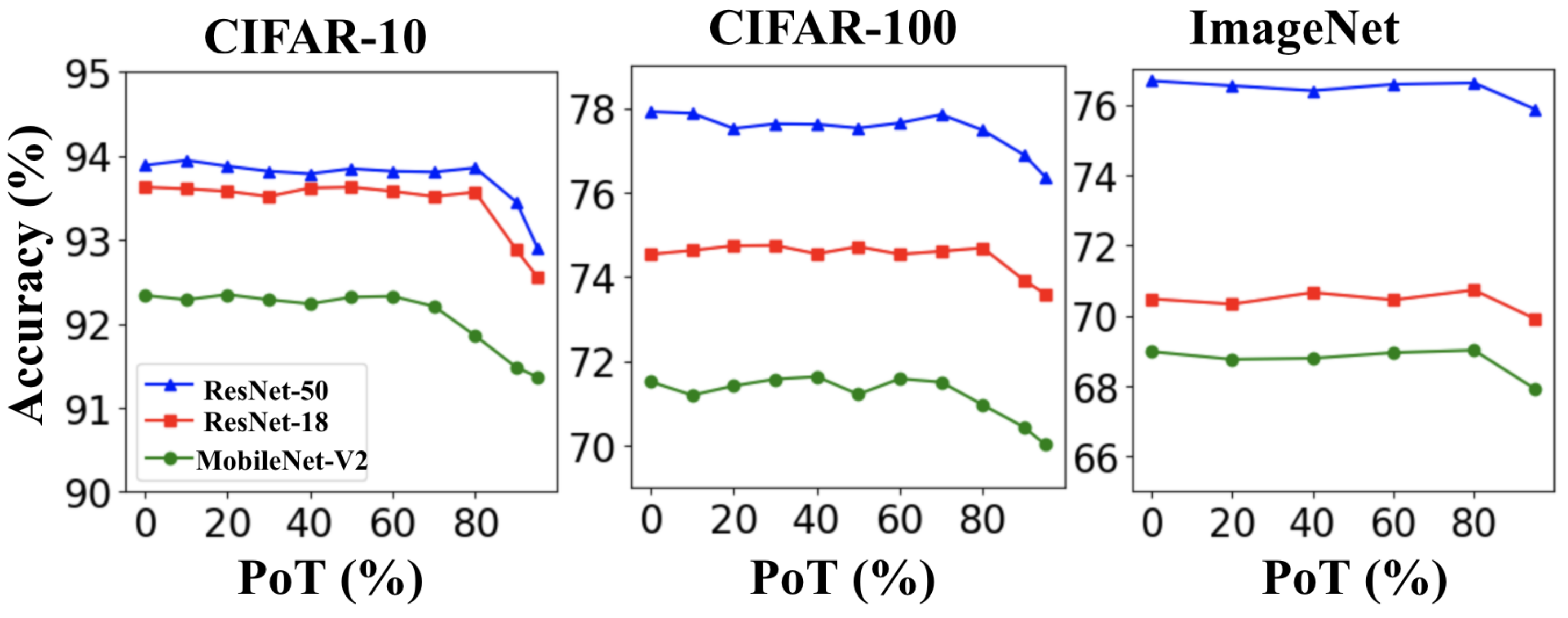}
\caption{\textbf{The effect of PoT-W4A4 ratio on the accuracy of ResNet-50, ResNet-18, and MobileNet-V2 models on CIFAR-10, CIFAR-100, and ImageNet datasets.} In the top row, only Fixed-W4A4 is mixed with PoT-W4A4. In the bottom row, 5\% of Fixed-W8A4 is used, besides PoT-W4A4 and Fixed-W4A4.}
\label{fig:PoTRatio}
\end{figure}

Figure \ref{fig:Precision} demonstrates the proposed multi-precision approach aligned with the row-wise mixed schemes.
The majority of the rows use the 4-bit precision for weights/activations i.e., PoT-W4A4 and Fixed-W4A4, because 2-bit has large accuracy loss and 3-bit is not suitable for hardware implementation, which prefers operands in 2-bit, 4-bit, 8-bit, etc.
To boost accuracy, a higher precision with 8-bit weights and 4-bit activations is used on the Fixed scheme, i.e., Fixed-W8A4.
The PoT scheme is not applied the higher precision because of its rigid resolution issue.
The ratio of PoT-W4A4 : Fixed-W4A4 : Fixed-W8A4 can be determined offline, and is the same across different layers, in order to keep the layer-wise uniformality in the hardware implementation, such that little overhead is incurred during the layer-by-layer inference execution and inference speedup can be guaranteed. 
In general, a larger portion of the PoT-W4A4 can help with inference speedup, but will degrade accuracy.
By using a small portion of Fixed-W8A4, the accuracy degradation due to the PoT scheme can be mitigated, as shown in Figure \ref{fig:PoTRatio}.
This work fixes the percentage of Fixed-W8A4 as 5\%, as it is enough to mitigate the accuracy degradation and an even smaller percentage only results in marginal further speedup of inference.

\subsection{RMSMP Quantization Algorithm}

\begin{algorithm}[tb]
\caption{RMSMP Quantization} \label{algo:MSQQuantization}
\small
\SetKwInOut{Input}{input}
\SetKwInOut{Output}{output}
\SetKwFunction{SGD}{SGD}
\Input{32-bit floating-point DNN model $\mathcal{M}$ with weights $W$;
Ratio ($\mathbb{R}$) of $\mathbb{S}_\text{PoT-4}: \mathbb{S}_\text{Fixed-4}: \mathbb{S}_\text{Fixed-8}$ \\
}
\Output{Quantized model $\hat{\mathcal{M}}$ }
$g \leftarrow \frac{\partial Loss}{\partial W}$;\\
\ForEach{\textnormal{layer} $i$}{
\ForEach{\textnormal{filter} $W_{ij}$ \textnormal{in} $i$}{
\tcp{power iteration}
Initialize random vector $v_0$; \\
\ForEach{$k\in\{1, \dots, max\_iter$\}}{  
Normalize $v_{k-1} \leftarrow \frac{v_{k-1}}{||v_{k-1}||_2}$; \\
$v_{k} \leftarrow \frac{\partial(g_{ij}^{\mathrm{T}}v_{k-1})}{\partial {W_{ij}}}$; \tcp{Eq.~\ref{solvehessian}} 
}
$\lambda_{W_{ij}} \leftarrow \max(v_{max\_iter})$; \\

}
\eIf{$\lambda_{W_{ij}}$ in top 5\%}{
$\mathbb{S}_{W_{ij}} \leftarrow \mathbb{S}_\text{Fixed-8}$;
}
{
Sort $W_{ij}$ by variance, get threshold $\theta$ by $\mathbb{R}$; \\

$\mathbb{S}_{W_{ij}} \leftarrow \mathbb{S}_\text{PoT-4}$ if variance lower than $\theta$;\\

$\mathbb{S}_{W_{ij}} \leftarrow \mathbb{S}_\text{Fixed-4}$ if variance higher than $\theta$; \\

}

}
\ForEach{epoch}{

\ForEach{batch}{
\tcp{forward propagation}
\ForEach{layer i}{
$out_i \leftarrow \textbf{proj}_{\mathbb{S}}(input_{i}) \cdot \textbf{proj}_{\mathbb{S}}(W_{i})$;
}
\tcp{back propagation}
$\frac{\partial Loss}{\partial W_i} \leftarrow \frac{\partial Loss}{\partial \textbf{proj}_{\mathbb{S}}(W_{i})}$; \tcp{STE Eq.~\ref{eqn:STE}}
}
}
Return $\hat{\mathcal{M}} \leftarrow \mathcal{M}\{\textbf{proj}_\mathbb{S}(W)\}$;

\end{algorithm}

\begin{table*}[t]
\small
\centering
\tabcolsep 6pt
\begin{tabular}{c c c c c c c}
\toprule
Quantization & \multicolumn{2}{c}{MobileNet-v2 Accuracy (\%)} & \multicolumn{2}{c}{ResNet-18 Accuracy (\%)} & \multicolumn{2}{c}{ResNet-50 Accuracy (\%)}\\
Method & Top-1 & Top-5 & Top-1 & Top-5 & Top-1 & Top-5\\
\hline
\multicolumn{7}{c}{\textbf{CIFAR-10}} \\
\hline
Baseline (W32A32)   & 92.51 & - & 93.62 & - & 93.91& -\\
Fixed-W4A4  &  91.76 (92.34)  & - & 92.9 (93.43) & - & 93.16 (93.73)\\
PoT-W4A4  & 90.92 (91.34) & - & 92.14 (92.97) & - & 92.53 (93.15)& -\\
APoT-W4A4 \cite{li2019additive}  &91.83 (92.72) & - & 92.94 (93.47) & - &92.87 (93.68) & -\\
PoT-W4A4 + Fixed-W4A4 & 91.38 (91.77)  & -  & 92.66 (93.21)   &-   & 93.14 (93.76) & -\\
APoT-W4A4 + Fixed-W4A4~\cite{chang2021mix}
& 91.99 (92.55) & - & 92.98 (93.65)  & - & 93.22 (93.77) & - \\
Fixed-W4A4 + Fixed-W8A4 & 92.54&-  & 93.69 & - &94.11 & - \\
\bf{RMSMP}  & \bf{92.58} & - & \bf{93.72}&- & \bf{94.03}& -\\
\hline
\multicolumn{7}{c}{\textbf{CIFAR-100}} \\
\hline
Baseline (W32A32)
         &  71.48& 91.98 & 74.49&  92.70 &  77.77 & 94.00\\
Fixed-W4A4
         & 70.22 (71.16) & 90.88 (91.63)
         & 73.88 (74.37) & 91.72 (92.31) 
         & 76.67 (77.41) & 93.22 (93.85)\\
PoT-W4A4
         & 67.11 (68.68) & 89.21 (90.06) 
         & 72.97 (73.88) & 91.65 (92.14) 
         & 74.58 (76.83) & 92.22 (92.74)\\
APoT-W4A4 \cite{li2019additive}   
         & 70.21 (71.13) & 90.85 (91.69)
         & 73.97 (74.33) & 92.03 (92.49) 
         & 76.58 (77.56) & 92.94 (93.55)\\
PoT-W4A4 + Fixed-W4A4
         & 68.94 (70.37)  & 90.05 (90.89)  
         & 73.41 (74.12)  & 91.68 (92.25)  
         & 76.95 (77.28) & 92.87 (93.66) \\
APoT-W4A4 + Fixed-W4A4~\cite{chang2021mix}
         & 70.25 (71.50) & 90.92 (91.82) 
         & 74.03 (74.60) & 92.05 (92.63)
         & 76.97 (77.31) & 92.99 (93.86)\\
Fixed-W4A4 + Fixed-W8A4  
         & 71.52 & 91.99 & 74.54 & 92.61 & 77.92 & 94.18\\
\bf{RMSMP}  & \bf{71.51} & \bf{91.97}
            & \bf{74.61} & \bf{92.69} & \bf{77.85} & \bf{94.13} \\
\hline
\multicolumn{7}{c}{\textbf{ImageNet}}  \\
\hline
Baseline (W32A32)
           & 71.88& 90.29 & 70.25 & 89.48
           & 76.51& 93.09\\
Fixed-W4A4      
           & 67.23 (69.26) & 86.03 (88.18)
           & 68.66 (69.72) & 87.54 (88.67) 
           & 75.58 (76.22) & 92.01 (92.43)\\
PoT-W4A4        
           & 65.88 (67.93) & 84.83 (86.63)
           & 67.11 (68.20) & 85.93 (87.14) 
           & 74.77 (75.03) & 91.66 (92.22)\\
APoT-W4A4~\cite{li2019additive}
           & 67.76 (68.97) & 86.42 (88.17)
           & 68.48 (70.70) & 87.92 (89.60) 
           & 75.49 (76.60) & 92.93 (93.10)\\
PoT-W4A4 + Fixed-W4A4
           & 66.20 (68.54)  & 85.66 (87.65)   
           & 67.98 (68.94)  & 86.75 (88.66)  
           & 75.85 (76.19) & 91.97 (92.77)\\
APoT-W4A4 + Fixed-W4A4~\cite{chang2021mix}
           & 67.31 (68.99) & 86.11 (88.04)
           & 69.22 (70.27) & 88.33 (89.42) 
           & 76.11 (76.22) & 92.58 (92.86) \\
Fixed-W4A4 + Fixed-W8A4
           & 68.98 & 89.04 & 70.48 & 89.52 & 76.68 & 93.44 \\
\bf{RMSMP} & \bf{69.02} & \bf{89.07}
           & \bf{70.73} & \bf{89.62}
           & \bf{76.62} & \bf{93.36}\\

\bottomrule
\end{tabular}
\caption{\textbf{Results from different quantization methods for MobileNet-v2, ResNet-18, and ResNet-50 models on CIFAR-10, CIFAR-100, and ImageNet datasets.} Baselines are unquantized models with 32-bit floating-point weights/activations i.e., W32A32. We explore different quantization schemes including Fixed-Point (Fixed), Power-of-Two (PoT), Additive Power-of-Two (APoT), and their combinations (PoT + Fixed with a ratio of 1:1, and APoT + Fixed with a ratio of 3:2). For these schemes, we report accuracy with 4-bit precision for weights/activations i.e., W4A4; and the accuracy in brackets is by relaxing the precision of the first and last layers to W8A8, while other layers are still in W4A4.
In the last two methods, row-wise multi-precision is introduced by (i) incorporating 4-bit Fixed with 8-bit Fixed (Fixed-W4A4 + Fixed-W8A4) with a ratio of 95:5, and (ii) incorporating PoT-W4A4, Fixed-W4A4 and Fixed-W8A4 (RMSMP), with a ratio of 65:30:5. In all the quantization methods with mixed schemes and multiple precisions, the combination is row-wise within a layer.}
\label{tab:comparison_schemes}
\end{table*}

The RMSMP quantization algorithm can train a DNN model from scratch or quantize a pre-trained model into a quantized one, such that for each layer, the numbers of filters quantized into PoT-W4A4, Fixed-W4A4, and Fixed-W8A4 follow the predefined ratio of $\mathbb{S}_\text{PoT-4}: \mathbb{S}_\text{Fixed-4}: \mathbb{S}_\text{Fixed-8}=A:B:C$, where $A+B+C=100$.

For the assignment of quantization schemes and precisions to the filters of each layer, we use the Hessian-based method to determine which filters should use Fixed-W8A4 (higher precision). And for the rest filters, we determine PoT-W4A4 vs Fixed-W4A4 based on the variances of the weights in each filter.
Once it is determined the assignment of quantization scheme and precision (PoT-W4A4, Fixed-W4A4, and Fixed-W8A4) down to the filter level for each layer, the Straight Through Estimator (STE) ~\cite{DBLP:journals/corr/BengioLC13,DBLP:journals/corr/abs-1903-05662}
\begin{equation}
\begin{aligned}
&\textbf{Forward}: y=\text{round}(x) \\
&\textbf{Backward}: \frac{\partial y}{\partial x}=\textbf{1}_{x\in R}
\end{aligned}
\label{eqn:STE}
\end{equation}
can be used to quantize the model into the specified schemes and precisions while solving the unavailable gradient issue during the backpropagation of the quantization process from continuous values into discrete values. 

In order to determine the filters with the higher precision (Fixed-W8A4), we extend from the Hessian based method in HAWQ~\cite{dong2019hawq} that {by computing the max} eigenvalue of the Hessian matrix of each filter, the filters with top 5\% eigenvalues get assigned the higher precision.
Hessian matrix is the second order derivatives of weight parameters, as following
\begin{equation}
H = \frac{\partial^2 Loss}{\partial W^2} = \frac{\partial g^\mathrm{T}}{\partial W}
\label{eq:hessianmatrix}
\end{equation}
where the first order gradient $g$ can be calculated through normal backpropagation process.

To solve the eigenvalues of Hessian matrix $H$ (i.e., $\lambda_H$) effectively, we employ the power iteration method as proposed in \cite{dong2019hawq, shen2020q}. 
By introducing an auxiliary variable $v$, we loop $v_{k+1}=H\cdot v_k$ as follows:
\begin{equation}
v_{k+1} = H\cdot v_k = \frac{\partial g^\mathrm{T}}{\partial W} \cdot v_k = \frac{\partial(g{^\mathrm{T}}v_k)}{\partial W}
\label{solvehessian}
\end{equation}
so that $\lim_{k\to \infty}\frac{v_{k+1}}{v_k} = \lambda_H$ and
 the last equality is proved in~\cite{dong2019hawq}.
With Eq. \ref{solvehessian}, we can proceed to solve $v_{k+1}$ given $v_k$ by computing the right term through simple backpropagation, thus solving $\lambda_H$.

Further, the scheme assignment is determined based on the variances of the filters. 
Finally, after the filters are assigned the quantization schemes and precisions, 
we can train our quantized model through STE. 
{Please note that we update the assignments of filters 
for every 10 epochs.}
The detailed steps are shown in Algorithm~\ref{algo:MSQQuantization}.
\section{Evaluation}\label{sec:eva}

\subsection{Experiment Setup}

\begin{table}[t]
\small
\centering
\tabcolsep 2.2pt
\begin{tabular}{ccccccc}
\toprule
\multirow{2}{*}{Method} & \multirow{2}{*}{{Approach}} &
\multirow{2}{*}{Bit-Width} & \multicolumn{2}{c}{First/Last} & Top-1 & Top-5 \\
~  & ~ & ~ & \multicolumn{2}{c}{Layer} & (\%) & (\%)\\
\hline
Baseline & - & W32A32 & $\times$ & $\times$ & 70.25 & 89.48\\
Dorefa~\cite{zhou2016dorefa} & Linear & W4A4 & $\times$ & $\times$ & 68.10 & 88.10 \\
PACT~\cite{choi2018pact} & Linear & W4A4 & $\times$ & $\times$ & 69.20 & 89.00 \\
DSQ~\cite{gong2019differentiable} & Linear & W4A4 & $\times$ & $\times$ & 69.56 & N/A \\
QIL~\cite{jung2019learning} & Linear & W4A4 & $\times$ & $\times$ & 70.10 & N/A \\
$\mu$L2Q~\cite{cheng2019uL2Q} & Linear & W4A4 & - & $\times$ & 65.92 & 86.72 \\
APoT~\cite{li2019additive} & Non-Lin. & W4A4 & 8bit & 8bit &70.70 & 89.60 \\
LQ-Nets~\cite{zhang2018lq} & Non-Lin. & W4A4&$\times$ & $\times$ & 69.30 & 88.80\\
DNAS~\cite{wu2018mixed} & MP-Lin.& Mixed& $\times$ & $\times$ & 70.64 & N/A\\
MPDNN~\cite{uhlich2019mixed} & MP-Lin. & Mixed &\checkmark & \checkmark & 70.08 & N/A\\
MSQ~\cite{chang2021mix} & MS & W4A4 & $\times$ & $\times$ & 70.27 & 89.42 \\
\textbf{RMSMP (Ours)}  & MP-MS & W4A4*&\textbf{\checkmark}&\textbf{\checkmark} &\textbf{70.73} & \textbf{89.62}\\

\bottomrule
\end{tabular}
\caption{\textbf{Comparisons with existing quantization works for ResNet-18 on ImageNet, using the (equivalent) 4-bit precision.} The covered quantization approaches include linear, non-linear (Non-Lin.), layer-wise multi-precision with linear (MP-Lin.), mixed-scheme (MS), and our row-wise multi-precision, mixed-scheme (MP-MS). The baseline is the unquantized model with 32-bit floating-point weights/activations, i.e. W32A32. For our method, W4A4* indicates that $5\%$ of the weights are in 8-bit (the activations are all in 4-bit). Of particular interest are the  "First/Last Layer" columns, which provide the detailed precision for the first and last layers in those work: \checkmark~means the same quantization as in other layers; $\times$ means no quantization (still 32-bit floating-point); and 8bit means 8-bit precision is used for the first/last layers.}
\label{tab:imagenetresnet}
\end{table}

\begin{table}[t]
\small
\centering
\tabcolsep 2.2pt
\begin{tabular}{ccccccc}
\toprule
\multirow{2}{*}{Method} & \multirow{2}{*}{Approach} &
\multirow{2}{*}{Bit-Width} & \multicolumn{2}{c}{First/Last} & Top-1 & Top-5 \\
~ & ~ & ~ & \multicolumn{2}{c}{Layer} & (\%) & (\%)\\
\hline
Baseline & - & W32A32& $\times$& $\times$ &76.51 & 93.09\\
Dorefa~\cite{zhou2016dorefa}  &Linear & W4A4&$\times$&$\times$ & 71.40 & 88.10\\
PACT~\cite{choi2018pact} & Linear & W4A4 & $\times$ & $\times$ & 76.50 & 93.30\\
APoT~\cite{li2019additive} & Non-Lin.& W4A4 & 8bit & 8bit &76.60 & 93.10\\
LQ-Nets~\cite{zhang2018lq} & Non-Lin.& W4A4&$\times$&$\times$& 75.40 & 92.40 \\
HAQ~\cite{wang2019haq} & MP-Lin. & Mixed &  8bit& \checkmark & 76.15 & 92.89 \\
MSQ~\cite{chang2021mix} & MS & W4A4 &$\times$ &$\times$ & 76.22 & 92.86\\
\textbf{RMSMP (Ours)}  & MP-MS & W4A4* & \textbf{\checkmark} & \textbf{\checkmark} & \textbf{76.62} & \textbf{93.36} \\

\bottomrule
\end{tabular}

\caption{\textbf{Comparisons with existing quantization work for ResNet-50 on ImageNet, using the (equivalent) 4-bit precision.}}
\label{tab:imagenetresnet50}
\end{table}

\begin{table}[t]
\small
\centering
\tabcolsep 2.2pt
\begin{tabular}{ccccccc}
\toprule
\multirow{2}{*}{Method} & \multirow{2}{*}{Approach} &
\multirow{2}{*}{Bit-Width} & \multicolumn{2}{c}{First/Last} & Top-1 & Top-5 \\
~  & ~ & ~ & \multicolumn{2}{c}{Layer} & (\%) & (\%)\\
\hline
Baseline  & - & W32A32& $\times$ &$\times$  &71.88 & 90.29\\

PACT~\cite{choi2018pact} & Linear & W4A4 & $\times$ & $\times$ & 61.40 & N/A\\
DSQ~\cite{gong2019differentiable} & Non-Lin.& W4A4&$\times$&$\times$& 64.80 & N/A\\
HAQ~\cite{wang2019haq} & MP-Lin. & Mixed &  8bit& \checkmark & 67.01 & 87.46 \\
MSQ~\cite{chang2021mix} & MS & W4A4 & $\times$ & $\times$ & 68.99 & 88.04 \\
\textbf{RMSMP (Ours)}  & MP-MS & W4A4* & \textbf{\checkmark} &\textbf{\checkmark} & \textbf{69.02} & \textbf{89.07}\\

\bottomrule
\end{tabular}
\caption{\textbf{Comparisons with existing work for MobileNet-V2 on ImageNet, using the (equivalent) 4-bit precision.}}
\label{tab:imagenetmobilenetv2}
\end{table}

\begin{table}[t]
\small
\centering
\tabcolsep 2.2pt
\begin{tabular}{ccccccc}
\toprule
\multirow{2}{*}{Method} & \multirow{2}{*}{Approach} &
\multirow{2}{*}{Bit-Width}&
Evaluation&
Result
\\
~ & ~ & ~ & Metric & ($\%$) \\
\toprule
\multicolumn{5}{c}{{\textbf{BERT on SST-2}}} \\

\hline
Baseline  & - & W32A32& Accuracy & 93.00  \\
Fixed & Linear&W4A4& Accuracy & 92.78\\
PoT & Non-Lin.& W4A4 & Accuracy & 92.43\\
PoT + Fixed & Mixed& W4A4 & Accuracy & 92.78\\
QBERT~\cite{shen2020q} & MP-Lin. & Mixed & Accuracy & 92.66   \\
\textbf{RMSMP (Ours)}  & MP-MS & W4A4* & Accuracy & \textbf{92.87}\\
\bottomrule
\multicolumn{5}{c}{{\textbf{BERT on MNLI}}}\\
\hline
Baseline  & - & W32A32& mm-Acc. & 84.90  \\
Fixed & Linear&W4A4& mm-Acc. & 84.71\\
PoT & Non-Lin.& W4A4 & mm-Acc. & 84.79\\
PoT + Fixed & Mixed& W4A4 & mm-Acc. & 84.51\\
QBERT~\cite{shen2020q} & MP-Lin. & Mixed & mm-Acc. & 84.17   \\
\textbf{RMSMP (ours)} & MP-MS & W4A4* & mm-Acc. & \textbf{84.83}\\
\bottomrule
\end{tabular}
\caption{\textbf{Comparisons with existing work for the BERT model on SST-2 and MNLI datasets, using the (equivalent) 4-bit precision.} The evaluation metric of SST-2 is accuracy, while the evaluation metric of MNLI is mismatched accuracy (mm-Acc.). For both metrics, a higher value is better.}
\label{tab:bert}
\end{table}

\begin{table*}[t]
\small
\centering
\tabcolsep 2.2pt
\begin{tabular}{c|c|c||c|c||c|c|c|c||c|c|c|c}
\toprule
\multirow{3}{*}{\makecell{Quantization \\ Method}} & \multirow{3}{*}{\makecell{PoT-W4A4 \\: Fixed-W4A4 \\: Fixed-W8A4}} & \multirow{3}{*}{\makecell{First/Last \\ Layer \\ Quantization}} & \multicolumn{2}{c||}{Accuracy} & \multicolumn{4}{c||}{Results on FPGA XC7Z020} & \multicolumn{4}{c}{Results on FPGA XC7Z045} \\
\cline{4-13}
 & & & Top-1 & Top-5 & \multicolumn{2}{c|}{Utilization} & Throughput & Latency & \multicolumn{2}{c|}{Utilization} & Throughput & Latency \\
\cline{6-7} \cline{10-11}
 & & & (\%) & (\%) & LUT & DSP & (GOP/s) & (ms) & LUT & DSP & (GOP/s) & (ms) \\
\hline
(1) Fixed & 0:100:0 & 8-bit Fixed & 69.72 & 88.67 & 26\% & 100\% & 29.6 & 122.6 & 21\% & 100\% & 115.6 & 31.4 \\
(2) Fixed & 0:100:0 & \checkmark & 68.66 & 87.54 & 23\% & 100\% & 36.5 & 99.3 & 19\% & 100\% & 142.7 & 25.4 \\
(3) PoT & 100:0:0 & 8-bit Fixed & 68.20 & 87.14 & 41\% & 100\% & 62.4 & 58.1 & 40\% & 100\% & 290.5 & 12.5 \\
(4) PoT & 100:0:0 & \checkmark & 67.11 & 85.93 & 43\% & 12\% & 72.2 & 50.2 & 43\% & 3\% & 352.6 & 10.3 \\
(5) PoT + Fixed & 50:50:0 & 8-bit Fixed & 68.94 & 88.66 & 50\% & 100\% & 50.3 & 72.0 & 48\% & 100\% & 196.8 & 18.4 \\
(6) PoT + Fixed & 50:50:0 & \checkmark & 67.98 & 86.75 & 46\% & 100\% & 75.8 & 47.8 & 45\% & 100\% & 296.3 & 12.2 \\
(7) PoT + Fixed & 60:40:0 & 8-bit Fixed & 68.53 & 88.47 & 52\% & 100\% & 57.0 & 63.6 & - & - & - & - \\
(8) PoT + Fixed & 67:33:0 & 8-bit Fixed & 68.46 & 88.22 & - & - & - & - & 63\% & 100\% & 245.8 & 14.8 \\

MSQ-1~\cite{chang2021mix}&60:40 &\checkmark & 69.22 & 88.33 & 53\% & 100\% & 77.0 & 47.1 & - & - & - & -\\
MSQ-2~\cite{chang2021mix} &67:33 &\checkmark & 69.14 & 88.17 & - & - & - & - & 66\% & 100\% & 359.2 & 10.1 \\
\bf{RMSMP-1} & 60:35:5 & \checkmark & \bf{70.66} & \bf{89.53} & 57\% & 100\% & \bf{89.0} & \bf{40.7} & - & - & - & - \\
\bf{RMSMP-2} & 65:30:5 & \checkmark & \bf{70.73} & \bf{89.62} & - & - & - & - & 67\% & 100\% & \bf{421.1} & \bf{8.6} \\

\bottomrule
\end{tabular}
\caption{\textbf{Implementations of different quantization methods on two FPGA boards for ResNet-18 on ImageNet, using the (equivalent) 4-bit precision.} For methods (1)$\sim$(8) and the two RMSMP, each quantization method corresponds to a ratio of PoT-W4A4 : Fixed-W4A4 : Fixed-W8A4 within each layer. The first and last layers are quantized into either 8-bit Fixed, or the same as other layers (denoted by \checkmark). 
The last two rows are by MSQ \cite{chang2021mix} with the ratios of APoT-W4A4 and Fixed-W4A4 as 60:40 and 67:33, respectively. Note that MSQ  uses APoT instead of PoT, and does not use multi-precision within layers. 
The proposed RMSMP method is compared with other methods in terms of accuracy as well as resource utilization rate, throughput, and latency on two FPGAs. For each FPGA board, the utilization ratios of look-up tables (LUT) and digital signal processing blocks (DSP) under each quantization method are provided in percentage. The utilization of LUT demonstrates the effectiveness of quantization method on inference speedup. For XC7Z020, the total numbers of LUTs and DSPs are 53.2K and 220, respectively. For XC7Z045, the total numbers of LUTs and DSPs are 218.6K and 900.}
\label{tab:ablation}
\end{table*}

Our RMSMP is evaluated on the image classification and natural language processing (NLP) applications. All the model training and quantization are conducted on NVIDIA TITAN RTX and GeForce RTX 2080 Ti GPUs, with CUDA~10.2 and PyTorch~1.5 frameworks on Ubuntu~18.04 operating system. The quantization algorithm utilizes the same data augmentation techniques as those used in training of baseline (32-bit floating-point) models. Training tricks such as step or cosine learning rate decay and weight regularization are also the same for training the baselines and in the quantization algorithm.

The evaluated models on image classification tasks include ResNet-18, ResNet-50~\cite{he2016deep} and MobileNet-v2~\cite{sandler2018mobilenetv2} on CIFAR-10, CIFAR-100~\cite{krizhevsky2009learning} and ImageNet ILSVRC-2012~\cite{krizhevsky2012imagenet} datasets.
Baseline models in 32-bit floating-point representation for CIFAR-10 and CIFAR-100 datasets are trained from scratch for $150$ epochs, then quantized for $150$ epochs.
For ImageNet dataset, pre-trained 32-bit floating-point models are quantized for $90$ epochs. 
The initial learning rates are $8e-3$ for CIFAR-10, $4e-3$ for CIFAR-100, and $1e-2$ for ImageNet.
For NLP tasks with BERT~\cite{devlin2019bert} models, we evaluate on the Stanford Sentiment Treebank (SST-2) and Multi-Genre Natural Language Inference (MNLI) datasets from the General Language Understanding Evaluation (GLUE)~\cite{wang2018glue} benchmark. The pre-trained BERT models are from HuggingFace Transformer~\cite{Wolf2019HuggingFacesTS}. Quantization and finetuning are simultaneously performed for 3 epochs with the initial learning rate of $2e-5$.
{Besides, for all models, the maximum iteration number of the power iteration method is set to 20. }

In addition to the model accuracy, we demonstrate the hardware efficiency (in terms of resource utilization, throughput, and latency) of the proposed RMSMP compared with other quantization methods through implementing the architectures with heterogeneous GEMM cores on FPGA devices, i.e., Zynq XC7Z020 and XC7Z045. 
Different ratios of quantization schemes/precisions are realized by adjusting the ratio among the processing element (PE) array sizes in the GEMM cores.
Our optimal RMSMP implementation utilizes three heterogeneous GEMM cores, i.e., \hw{GEMM_{PoT-4}} for processing with PoT-W4A4 filters, \hw{GEMM_{Fixed-4}} for Fixed-W4A4, and \hw{GEMM_{Fixed-8}} for Fixed-W8A4.
The working frequency is set to 100MHz for all implementations.

\subsection{Accuracy Performance}

Table~\ref{tab:comparison_schemes} compares the proposed RMSMP method with other quantization methods for three models on three datasets. RMSMP generally achieves $0.5\% \sim 1.5\%$ higher top-1 accuracy than methods with Fixed, PoT, APoT, PoT + Fixed, and APoT + Fixed using 4-bit precision, and comparable accuracy to the Fixed-W4A4 + Fixed-W8A4 method.

The results in Tables~\ref{tab:imagenetresnet},~\ref{tab:imagenetresnet50}, and~\ref{tab:imagenetmobilenetv2} demonstrate the effectiveness of the proposed RMSMP method for three models i.e., ResNet-18, ResNet-50, and MobileNet-V2, respectively, on ImageNet using mainly 4-bit precision. For ResNet-18, RMSMP increases the top-1 accuracy by $0.48\%$ compared with the baseline, and outperforms existing methods by up to $4.81\%$. For ResNet-50, RMSMP increases the baseline accuracy by $0.11\%$, and outperforms existing methods by up to $5.22\%$.
As for MobileNet-V2, RMSMP achieves the least accuracy loss, with up to  $7.62\%$ increase on top-1 accuracy than other methods.
In addition, comparisons on a large language model BERT are provided in Table~\ref{tab:bert}. 
The BERT is a large model with higher redundancy. And therefore it is easy to achieve negligible accuracy loss after quantization.
All the quantization methods in Table~\ref{tab:bert} perform very well and RMSMP yields slightly better accuracy than other methods on SST-2 and MNLI datasets.

\subsection{Hardware Efficiency}
\label{sec:hardware_efficiency}

The proposed RMSMP method is compared with other quantization methods on two FPGA boards with the ResNet-18 model on the ImageNet dataset, as displayed in Table~\ref{tab:ablation}.
The total amount of resources on a specific FPGA board determines the optimal ratio of PoT-W4A4 : Fixed-W4A4 : Fixed-W8A4 schemes.
Specifically, the optimal ratio on XC7Z020 is 60:35:5 (RMSMP-1), resulting in top-1 accuracy of 70.66\% and latency of 40.7ms, while the optimal ratio on XC7Z045 is 65:30:5 (RMSMP-2), leading to top-1 accuracy of 70.73\% and latency of 8.6ms.
RMSMP with the optimal ratio of quantization schemes achieves up to $3.01\times$ speedup on XC7Z020 and up to $3.65\times$ speedup on XC7Z045, comparing with the single-scheme and single-precision method (1) Fixed.
Methods (1)(3)(5)(7)(8) employ 8-bit Fixed quantization for the first and last layers to maintain the accuracy, whereas this hurts the inference speed. Compared with methods (1)(2)(5)(6), RMSMP enhances the utilization ratio of LUTs to increase the throughput. Additionally, RMSMP achieves higher accuracy and lower latency with full DSP usage compared with the method (4).
We also compare with MSQ \cite{chang2021mix} with the first/last layer quantized into 4-bit. Our RMSMP outperforms MSQ in both accuracy and inference speed. 
\section{Conclusion}\label{sec:con}

This work proposes a novel DNN quantization framework, namely RMSMP, with a {R}ow-wise {M}ixed-{S}cheme and {M}ulti-{P}recision approach, featuring (i) layer-wise uniformality to fulfill the requirement of practical hardware implementation, (ii) row-wise flexibility for mixed schemes and multiple precisions, (iii) hardware-informative selection of candidate schemes and precisions (bit-widths) for significantly reducing the algorithm search space, and (iv) best accuracy performance among the state-of-the-arts.
This paper adopts a highly hardware-informative solution strategy to reduce the problem search space. Evaluating with the image classification and natural language processing applications, our RMSMP achieves superior accuracy performance.
And we also evaluate our RMSMP with FPGA devices.


\section*{Acknowledgment}

{This work is partly supported by the National Science Foundation CCF-1901378, and CCF-1937500. Any opinions, findings, and conclusions or recommendations  in this material are those of the authors and do not necessarily reflect the views of NSF.}

{\small
\bibliographystyle{ieee_fullname}
\bibliography{egbib}

\begin{thebibliography}{10}\itemsep=-1pt

\bibitem{DBLP:journals/corr/BengioLC13}
Yoshua Bengio, Nicholas L{\'e}onard, and Aaron Courville.
\newblock Estimating or propagating gradients through stochastic neurons for
  conditional computation.
\newblock {\em arXiv preprint arXiv:1308.3432}, 2013.

\bibitem{chang2021mix}
Sung-En Chang, Yanyu Li, Mengshu Sun, Runbin Shi, Hayden K-H So, Xuehai Qian,
  Yanzhi Wang, and Xue Lin.
\newblock Mix and match: A novel fpga-centric deep neural network quantization
  framework.
\newblock In {\em Proceedings of the 2021 IEEE International Symposium on High
  Performance Computer Architecture}, 2021.

\bibitem{cheng2019uL2Q}
Gong Cheng, Lu Ye, Li Tao, Zhang Xiaofan, Hao Cong, Chen Deming, and Chen Yao.
\newblock $\mu$l2q: An ultra-low loss quantization method for dnn.
\newblock {\em The 2019 International Joint Conference on Neural Networks
  (IJCNN)}, 2019.

\bibitem{choi2018pact}
Jungwook Choi, Zhuo Wang, Swagath Venkataramani, Pierce I-Jen Chuang,
  Vijayalakshmi Srinivasan, and Kailash Gopalakrishnan.
\newblock Pact: Parameterized clipping activation for quantized neural
  networks.
\newblock {\em arXiv preprint arXiv:1805.06085}, 2018.

\bibitem{courbariaux2015binaryconnect}
Matthieu Courbariaux, Yoshua Bengio, and Jean-Pierre David.
\newblock Binaryconnect: Training deep neural networks with binary weights
  during propagations.
\newblock In {\em Advances in neural information processing systems (NeurIPS)},
  pages 3123--3131, 2015.

\bibitem{courbariaux2016binarized}
Matthieu Courbariaux, Itay Hubara, Daniel Soudry, Ran El-Yaniv, and Yoshua
  Bengio.
\newblock Binarized neural networks: Training deep neural networks with weights
  and activations constrained to+ 1 or-1.
\newblock {\em arXiv preprint arXiv:1602.02830}, 2016.

\bibitem{devlin2018bert}
Jacob Devlin, Ming-Wei Chang, Kenton Lee, and Kristina Toutanova.
\newblock Bert: Pre-training of deep bidirectional transformers for language
  understanding.
\newblock {\em arXiv preprint arXiv:1810.04805}, 2018.

\bibitem{devlin2019bert}
Jacob Devlin, Ming-Wei Chang, Kenton Lee, and Kristina Toutanova.
\newblock Bert: Pre-training of deep bidirectional transformers for language
  understanding.
\newblock In {\em Proceedings of the 2019 Conference of the North American
  Chapter of the Association for Computational Linguistics: Human Language
  Technologies, Volume 1 (Long and Short Papers)}, pages 4171--4186, 2019.

\bibitem{dong2019hawqv2}
Zhen Dong, Zhewei Yao, Yaohui Cai, Daiyaan Arfeen, Amir Gholami, Michael~W
  Mahoney, and Kurt Keutzer.
\newblock Hawq-v2: Hessian aware trace-weighted quantization of neural
  networks.
\newblock {\em arXiv preprint arXiv:1911.03852}, 2019.

\bibitem{dong2019hawq}
Zhen Dong, Zhewei Yao, Amir Gholami, Michael~W Mahoney, and Kurt Keutzer.
\newblock Hawq: Hessian aware quantization of neural networks with
  mixed-precision.
\newblock In {\em Proceedings of the IEEE International Conference on Computer
  Vision (ICCV)}, pages 293--302, 2019.

\bibitem{esser2019learned}
Steven~K Esser, Jeffrey~L McKinstry, Deepika Bablani, Rathinakumar Appuswamy,
  and Dharmendra~S Modha.
\newblock Learned step size quantization.
\newblock {\em International Conference on Learning Representations (ICLR)},
  2019.

\bibitem{gong2019differentiable}
Ruihao Gong, Xianglong Liu, Shenghu Jiang, Tianxiang Li, Peng Hu, Jiazhen Lin,
  Fengwei Yu, and Junjie Yan.
\newblock Differentiable soft quantization: Bridging full-precision and low-bit
  neural networks.
\newblock In {\em Proceedings of the IEEE International Conference on Computer
  Vision (ICCV)}, pages 4852--4861, 2019.

\bibitem{han2015learning}
Song Han, Jeff Pool, John Tran, and William Dally.
\newblock Learning both weights and connections for efficient neural network.
\newblock In {\em Advances in neural information processing systems (NeurIPS)},
  pages 1135--1143, 2015.

\bibitem{he2016deep}
Kaiming He, Xiangyu Zhang, Shaoqing Ren, and Jian Sun.
\newblock Deep residual learning for image recognition.
\newblock In {\em Proceedings of the IEEE conference on computer vision and
  pattern recognition (CVPR)}, pages 770--778, 2016.

\bibitem{he2019simultaneously}
Zhezhi He and Deliang Fan.
\newblock Simultaneously optimizing weight and quantizer of ternary neural
  network using truncated gaussian approximation.
\newblock In {\em Proceedings of the IEEE Conference on Computer Vision and
  Pattern Recognition (CVPR)}, pages 11438--11446, 2019.

\bibitem{jung2019learning}
Sangil Jung, Changyong Son, Seohyung Lee, Jinwoo Son, Jae-Joon Han, Youngjun
  Kwak, Sung~Ju Hwang, and Changkyu Choi.
\newblock Learning to quantize deep networks by optimizing quantization
  intervals with task loss.
\newblock In {\em Proceedings of the IEEE Conference on Computer Vision and
  Pattern Recognition (CVPR)}, pages 4350--4359, 2019.

\bibitem{krizhevsky2009learning}
Alex Krizhevsky.
\newblock Learning multiple layers of features from tiny images.
\newblock 2009.

\bibitem{krizhevsky2012imagenet}
Alex Krizhevsky, Ilya Sutskever, and Geoffrey~E Hinton.
\newblock Imagenet classification with deep convolutional neural networks.
\newblock In {\em Advances in neural information processing systems}, pages
  1097--1105, 2012.

\bibitem{leng2018extremely}
Cong Leng, Zesheng Dou, Hao Li, Shenghuo Zhu, and Rong Jin.
\newblock Extremely low bit neural network: Squeeze the last bit out with admm.
\newblock In {\em Thirty-Second AAAI Conference on Artificial Intelligence
  (AAAI)}, 2018.

\bibitem{li2016ternary}
Fengfu Li, Bo Zhang, and Bin Liu.
\newblock Ternary weight networks.
\newblock {\em arXiv preprint arXiv:1605.04711}, 2016.

\bibitem{li2019additive}
Yuhang Li, Xin Dong, and Wei Wang.
\newblock Additive powers-of-two quantization: An efficient non-uniform
  discretization for neural networks.
\newblock In {\em International Conference on Learning Representations (ICLR)},
  2020.

\bibitem{lin2017towards}
Xiaofan Lin, Cong Zhao, and Wei Pan.
\newblock Towards accurate binary convolutional neural network.
\newblock In {\em Advances in Neural Information Processing Systems (NeurIPS)},
  pages 345--353, 2017.

\bibitem{lou2019autoq}
Qian Lou, Feng Guo, Minje Kim, Lantao Liu, and Lei Jiang.
\newblock Autoq: Automated kernel-wise neural network quantization.
\newblock In {\em International Conference on Learning Representations (ICLR)},
  2019.

\bibitem{DBLP:journals/corr/MiyashitaLM16}
Daisuke Miyashita, Edward~H Lee, and Boris Murmann.
\newblock Convolutional neural networks using logarithmic data representation.
\newblock {\em arXiv preprint arXiv:1603.01025}, 2016.

\bibitem{park2018value}
Eunhyeok Park, Sungjoo Yoo, and Peter Vajda.
\newblock Value-aware quantization for training and inference of neural
  networks.
\newblock In {\em Proceedings of the European Conference on Computer Vision
  (ECCV)}, pages 580--595, 2018.

\bibitem{rastegari2016xnor}
Mohammad Rastegari, Vicente Ordonez, Joseph Redmon, and Ali Farhadi.
\newblock Xnor-net: Imagenet classification using binary convolutional neural
  networks.
\newblock In {\em European conference on computer vision (ECCV)}, pages
  525--542. Springer, 2016.

\bibitem{sandler2018mobilenetv2}
Mark Sandler, Andrew Howard, Menglong Zhu, Andrey Zhmoginov, and Liang-Chieh
  Chen.
\newblock Mobilenetv2: Inverted residuals and linear bottlenecks.
\newblock In {\em Proceedings of the IEEE conference on computer vision and
  pattern recognition (CVPR)}, pages 4510--4520, 2018.

\bibitem{shen2020q}
Sheng Shen, Zhen Dong, Jiayu Ye, Linjian Ma, Zhewei Yao, Amir Gholami,
  Michael~W Mahoney, and Kurt Keutzer.
\newblock Q-bert: Hessian based ultra low precision quantization of bert.
\newblock In {\em Thirty-Second AAAI Conference on Artificial Intelligence
  (AAAI)}, pages 8815--8821, 2020.

\bibitem{uhlich2019mixed}
Stefan Uhlich, Lukas Mauch, Fabien Cardinaux, Kazuki Yoshiyama, Javier~Alonso
  Garcia, Stephen Tiedemann, Thomas Kemp, and Akira Nakamura.
\newblock Mixed precision dnns: All you need is a good parametrization.
\newblock {\em International Conference on Learning Representations (ICLR)},
  2020.

\bibitem{wang2018glue}
Alex Wang, Amanpreet Singh, Julian Michael, Felix Hill, Omer Levy, and Samuel
  Bowman.
\newblock Glue: A multi-task benchmark and analysis platform for natural
  language understanding.
\newblock In {\em Proceedings of the 2018 EMNLP Workshop BlackboxNLP: Analyzing
  and Interpreting Neural Networks for NLP}, pages 353--355, 2018.

\bibitem{wang2019haq}
Kuan Wang, Zhijian Liu, Yujun Lin, Ji Lin, and Song Han.
\newblock Haq: Hardware-aware automated quantization with mixed precision.
\newblock {\em International Conference on Computer Vision and Pattern
  Recognition (CVPR)}, 2019.

\bibitem{Wolf2019HuggingFacesTS}
Thomas Wolf, Lysandre Debut, Victor Sanh, Julien Chaumond, Clement Delangue,
  Anthony Moi, Pierric Cistac, Tim Rault, Rémi Louf, Morgan Funtowicz, Joe
  Davison, Sam Shleifer, Patrick von Platen, Clara Ma, Yacine Jernite, Julien
  Plu, Canwen Xu, Teven~Le Scao, Sylvain Gugger, Mariama Drame, Quentin Lhoest,
  and Alexander~M. Rush.
\newblock Huggingface's transformers: State-of-the-art natural language
  processing.
\newblock {\em ArXiv}, abs/1910.03771, 2019.

\bibitem{wu2018mixed}
Bichen Wu, Yanghan Wang, Peizhao Zhang, Yuandong Tian, Peter Vajda, and Kurt
  Keutzer.
\newblock Mixed precision quantization of convnets via differentiable neural
  architecture search.
\newblock {\em arXiv preprint arXiv:1812.00090}, 2018.

\bibitem{yao2019pyhessian}
Zhewei Yao, Amir Gholami, Kurt Keutzer, and Michael Mahoney.
\newblock Pyhessian: Neural networks through the lens of the hessian.
\newblock {\em arXiv preprint arXiv:1912.07145}, 2019.

\bibitem{DBLP:journals/corr/abs-1903-05662}
Penghang Yin, Jiancheng Lyu, Shuai Zhang, Stanley Osher, Yingyong Qi, and Jack
  Xin.
\newblock Understanding straight-through estimator in training activation
  quantized neural nets.
\newblock In {\em International Conference on Learning Representations (ICLR)},
  2018.

\bibitem{zhang2018lq}
Dongqing Zhang, Jiaolong Yang, Dongqiangzi Ye, and Gang Hua.
\newblock Lq-nets: Learned quantization for highly accurate and compact deep
  neural networks.
\newblock In {\em Proceedings of the European conference on computer vision
  (ECCV)}, pages 365--382, 2018.

\bibitem{DBLP:journals/corr/ZhouYGXC17}
Aojun Zhou, Anbang Yao, Yiwen Guo, Lin Xu, and Yurong Chen.
\newblock Incremental network quantization: Towards lossless cnns with
  low-precision weights.
\newblock In {\em International Conference on Learning Representations (ICLR)},
  2017.

\bibitem{zhou2016dorefa}
Shuchang Zhou, Yuxin Wu, Zekun Ni, Xinyu Zhou, He Wen, and Yuheng Zou.
\newblock Dorefa-net: Training low bitwidth convolutional neural networks with
  low bitwidth gradients.
\newblock {\em arXiv preprint arXiv:1606.06160}, 2016.

\bibitem{zhu2016trained}
Chenzhuo Zhu, Song Han, Huizi Mao, and William~J Dally.
\newblock Trained ternary quantization.
\newblock In {\em International Conference on Learning Representations (ICLR)},
  2017.

\end{thebibliography}
}

\end{document}